\documentclass[10pt,twocolumn,letterpaper]{article}
\pdfoutput=1

\usepackage{cvpr}
\usepackage{times}
\usepackage{epsfig}
\usepackage{graphicx}
\usepackage{amsmath}
\usepackage{amssymb}

\usepackage{hyperref}

\cvprfinalcopy 

\def\cvprPaperID{****} 

\begin{document}

\title{Improving precision and recall of face recognition in SIPP with combination of modified mean search and LSH}

\author{Xihua.Li\\
{\tt\small lixihua9@126.com}
}

\maketitle

\begin{abstract}
Although face recognition has been improved much as the development of Deep Neural Networks, SIPP(Single Image Per Person) problem in face recognition has not been better solved, especially in practical applications where searching over complicated database. In this paper, a combination of modified mean search and LSH method would be introduced orderly to improve the precision and recall of SIPP face recognition without retrain of the DNN model. First, a modified SVD based augmentation method would be introduced to get more intra-class variations even for person with only one image. Second, an unique rule based combination of modified mean search and LSH method was proposed the first time to help get the most similar personID in a complicated dataset, and some theoretical explaining followed. Third, we would like to emphasize, no need to retrain of the DNN model and would easy to be extended without much efforts. We do some practical testing in competition of Msceleb challenge-2 2017 which was hold by Microsoft Research, great improvement of coverage from 13.39\% to 19.25\%, 29.94\%, 42.11\%, 47.52\% at precision 99\%(P99) would be shown latter, coverage reach 94.2\% and 100\% at precision 97\%(P97) and 95\%(P95) respectively. As far as we known, this is the only paper who do not fine-tuning on competition dataset and ranked top-10. A similar test on CASIA WebFace dataset also demonstrated the same improvements on both precision and recall.
\end{abstract}

\section{Introduction}

Since 2012, Deep Neural Networks has been utilized in almost every aspects of computer vision, such as image classification and recognition, object detection, tracking and recognition, OCR in natural images, face recognition, saliency detection, 3D action recognition, image segmentation, super-resolution, image creation, content-base-image-retrieval, medical image diagnosis, and so on. At the same time, deep neural networks itself has also improved much, from AlexNet to GoogleNet, VGG, ResNet, Inception, Inception-ResNet\cite{DBLP:journals/corr/SzegedyIV16}, which focus on mainly two aspects. First, to reduce parameters and computing of deep neural networks. Second, to improve the feature representation ability of deep neural networks. The destination is to find a neural networks which has the best representation of what's in the picture with less computing.

Face Recognition is almost the most hot topic in computer vision and has now been put into use in many practical applications. At early stage, Fisher face and Eigen face\cite{doi:10.1162/jocn.1991.3.1.71} were used to regard face recognition as a problem of finding a suitable projection that can mostly maximum inter-class variations and minimum intra-class variations. In middle stage, face landmarks were detected and local features such as LBP are extracted to have a better representation of face with contextual information included. Nowadays, almost every top face recognition algorithms are deep neural networks based. With different kinds of loss function designed, deep neural network based face recognition methods has improved much and now 99.8\% precision has been reached on LFW benchmark which has already exceed human beings\cite{10.5120/ijais2016451597}. The development of face recognition itself also following two trends above: better representation ability and less computing. Better representation ability in face recognition mainly means the ability to handle pose and expressions variations.

Different kinds of loss function is also a way for how us human beings to treat what kind of machine learning problem face recognition is. At the beginning, face recognition was formulated as a classification problem, but when face IDs increase sharply, classification precision decreased. Then, verification based loss function occur, deep neural networks was used to learn a representation that could verification whether the two faces belongs to a same person or not. DeepID, DeepID2, DeepID3\cite{DBLP:journals/corr/SunLWT15} perhaps is the most successful face recognition algorithms that profit from the design of loss function.

Different kinds of networks and loss function were essentially used to learn a powerful representation of human face with lower intra-class variations and higher inter-class variations. But if we already learnt this representation, is there any way that we could do to improve the face recognition precision and recall without much efforts? Under this consideration, large scale face recognition problem is a way for us to search for the most nearest points in high dimensions feature space when given a particular face feature vector. In this paper, we take Msceleb challenge-2 2017 hold by Microsoft Research as background, a comprehensive evaluation on Msceleb dataset and CASIA WebFace dataset\footnote{http://www.cbsr.ia.ac.cn/english/CASIA-WebFace-Database.html}\cite{2014arXiv1411.7923Y} demonstrated the same conclusion.

As promising results shown in table1 and table3, we have 4 contributions in this paper.

\noindent \emph {1. A modified face augmentation method based on SVD was proposed which is easy to generating more augmented faces and get richer intra-class variations than ever.}

\noindent \emph {2. An unique rule based combination method of brute-force search and modified mean search was proposed the first time which increase precision and recall obviously.}

\noindent \emph {3. LSH was used to replace of brute-force search above which is more efficient and robust to noise.}

\noindent \emph {4. No need to retrain or fine-tuning of the DNN model which is different from most papers so far.}

\section{Msceleb challenge-2 2017 and Related works}

\subsection{Msceleb challenge-2 2017}

Msceleb face recognition challenge\cite{one-shot-face-recognition-promoting-underrepresented-classes}\cite{ms-celeb-1m-dataset-benchmark-large-scale-face-recognition-2} was hold by Microsoft Research which was named the "World Cup" for face recognition, attracted many researchers\footnote {http://www.msceleb.org/}. Challenge-2 was newly started, mainly focus on SIPP problem and also emphasize the generalization ability of the face recognition algorithm, which is almost the real-world scenarios. Challenge-2 was called Low-Shot Learning or Know you at One Glance. Here we have a brief introduction.

In challenge-2, we investigate the problem of low-shot face recognition, with the goal to build a large-scale face recognizer capable of recognizing a substantial number of individuals with high precision and recall. We create a benchmark dataset consisting of 21,000 persons each with 50-100 images of high accuracy ($>$99\%). We divide this dataset into the following two sets: \textbf{Baseset}, there are 20,000 persons in the baseset. Each person has 50-100 images for training, and about 5 images for testing. \textbf{Novelset}, there are 1,000 persons in the novelset. Each person has only 1 image for training, and 20 images for testing.

Our goal is to study when tens of images are given for each person in the baseset while only one image are given for each person in the novelset, how to develop an algorithm to recognize the persons in the both dataset.

Our measurement set contains a mixture of test images from both the baseset and the novelset. We mainly focus on the classification performance with the test images in the novelset to evaluate how well the computer can learn novel visual concepts with limited number of training samples, while also monitor the performance on the baseset to ensure that the performance gain on the novelset which is not obtained by sacrificing the performance on the baseset. A contesting system is asked to produce at least one prediction label with a confidence score per test image. To match with real scenarios, we measure the recognition coverage at a given precision 99\%. That is, for N images in the measurement set, if an algorithm recognizes M images, among which C images are correct, we will calculate precision and coverage(or recall) as:
\begin{equation}
precision = C/M
\end{equation}
\begin{equation}
coverage = M/N
\end{equation}

By varying the recognition confidence threshold, we can determine the coverage when the precision is at 99\%. We rank the methods according to the coverage at the 99\% precision with the test images in the novelset, while also monitor the performance on the baseset.

Intuitively, for person in baseset, it's easy to be recognized because it would be easy for us to find a personID whose pose and expressions same with search face. But it would be very hard for person in novelset, it would be easy to be confused for same pose and expressions but came from different personID, as in Figure 1.

\begin{figure}[t]
\center{\includegraphics[width=7cm]  {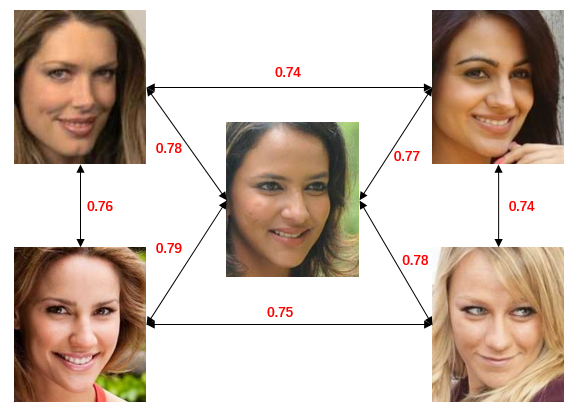}}
\caption{Confused similarity scores because of same pose and expressions. In fact, 5 faces belongs to 5 different persons and the surrounding 4 pictures are top-4 similarity pictures in dataset.}
\end{figure}

\subsection{Related works}
In fact, applications in Msceleb challenge-2 2017 is not a simple SIPP problem, its more complicated because of the interference of baseset. Yue Wu proposed a combination of Convolutional Neural Network and Nearest Neighbor method in \cite{Wu_2017_ICCV}, but the combination was done by a simple hybrid classifier. A GAN based data augmentation method was proposed in \cite{Choe_2017_ICCV} by Choe and transfer learning utilized. Doppelganger Mining for a better face representations was proposed by Smirnov in \cite{Smirnov_2017_ICCV}, which main contribution is to maintain a list with most similar identities for generating better mini-batches by sampling pairs of similar-looking identities. Yan Xu proposed Multi-Cognition Softmax Model(MCSM) in \cite{Xu_2017_ICCV} to distribute training data to several cognition units by a data shuffle strategy.

Content-based image similarity search is a difficult problem due to the high dimensionality and usually massive amount of image data. The main challenge is to achieve high-quality similarity search with high speed and low space usage\cite{Lv:2006:SSL:1237003}. Face recognition in complicated dataset is essentially a CBIR problem, but its totally different to a normal CBIR image search engines like Google, Bing and Flickr\cite{szHucs2017content}, we are not aimed at searching for nearest neighbor, but to find the particular personID. K-Means Clustering\cite{rejito2012optimization}, KD-tree\cite{kakde2005range}, Hashing\cite{wang2014hashing} are 3 most used searching methods in industry with the same idea of clustering samples in database and accelerating searching. On top of Hashing, local sensitive hashing has also been utilized in many methods\cite{datar2004locality}\cite{erin2015deep} which implied noise patience on some extent.

In this paper, we took CNN+NN method\cite{Wu_2017_ICCV} and LSH\cite{erin2015deep} as reference. A combination of modified mean search and LSH method was proposed.

\section{Face augmentation and Baseline}
\subsection{Modified SVD based face augmentation}
In this section, we proposed a modified SVD based method to do face augmentation for novelset faces. By using SVD\cite {GAO2008726}, we decompose the face image into two complementary parts: the first part in constructed by the SVD basis images associated with several largest singular values, and the second part is constructed by the other low-energy basis images. This first part preserves most of the energy of an image and reflects the general appearance of the image. The second part is the difference between the original image and the first part, it can reflect, to some extent, the variations of the same class face images, such as pose and expressions.

Given an aligned face image $ A\in R^{m*n} $ and suppose $ m\geqslant n $, we have the following expression according to the definition of SVD.
\begin{equation}
A=\sum_{i=1}^{n} \sigma _{i}\cdot \mu _{i}\cdot \upsilon _{i}
\end{equation}
where $ \mu _{i} $ are $ i $th column of $ U\in R^{m*m} $ and $ V\in R^{n*n} $, respectively. U and V are composed of the eigenvectors of $ AA^{T} $ and $ A^{T}A $, respectively, $ \sigma _{i}$ is the singular values of image A and we let $ \sigma _{1}\geqslant \sigma _{2}\geqslant ... \geqslant \sigma _{n}$.

\begin{equation}
Faces_{i,j,k}=merge(R_i, G_j, B_k)
\end{equation}
where i, j, k = 1.0, 0.95, 0.90, 0.85 and totaly 64 faces generated.

\begin{figure}[t]
\center{\includegraphics[width=6cm]  {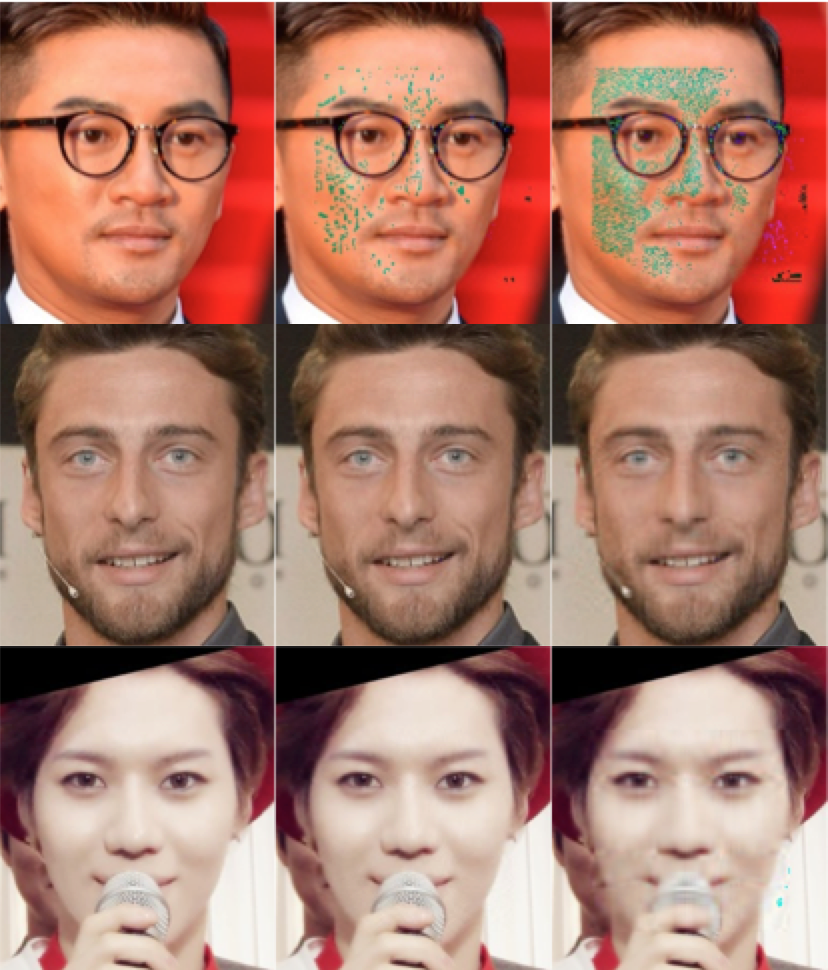}}
\caption{Examples of SVD augmented faces. Left to right: 100\%, 95\%, 90\% percentage of energy reserved. Top to bottom: noise added, blur, noise plus blur effects occurs for different faces.}
\end{figure}

For normal scenery, SVD based face augmentation was conducted on gray image and fewer images could get for reserving different proportion of energy. In our experiment, we try to reserve 95\%, 90\%, 85\%, 80\%, 75\% energy of the eigenvalues at first. After some more evaluations, we reserve 95\%, 90\%, 85\% which is a balance between degree of variations and image quality. \textbf {As equation 4, here we do SVD augmentation on each of RGB channel and merge them up which aimed at obtain more variations, in this way, one face image become 4*4*4=64 in total(3 channels and each has 4 energy choice)}. With same or different percentage of energy reserving for each channel, more variations would get compared with formal methods. Augmented face images could be seen in figure 2. In this way, we expand feature vector point of a specialized face to a sphere surrounding the point, like small blue circle in the right part of Figure 3. Note that, because of different pose and expression, the surrounding sphere may not in the middle of feature space of a particular person.

\subsection{SVD based method evaluation}
Here we have brief introduction about the baseline and evaluations methods. First, a Dlib\footnote {http://vis-www.cs.umass.edu/lfw/results.html} based face detection, face alignment and face feature extractor method were used in whole of the evaluation. Dlib deep neural network face recognition tools has a 99.38\% accuracy on the standard LFW face recognition benchmark, which is comparable to other state-of-the-art methods for face recognition until early of 2017. It's essentially a version of the ResNet-34 network from the paper Deep Residual Learning for Image Recognition by He, Zhang, Ren, and Sun with a few layers removed and the number of filters per layer reduced by half. The network was trained from scratch on a dataset of about 3 million faces. Second, for baseset there are 20,000 persons and each has 50-100 images, for novelset there are 1,000 persons each has only one image. We do face detection, alignment and face feature extraction on baseset and novelset and finally got 1,169,166 face feature vectors for 1,169,166 face images which belongs to 21,000 persons. Note please, for novelset, we do SVD on each face images and got 63 degrade face images for correspondence person, we also extract feature vectors on degraded face images with dilb face feature extractor, and more 63,000(=63*1,000) face feature vectors we got which totally became 1,232,166 face feature vectors. Notations, $ X _{i, j} $ is baseset feature vector where $ i\in [1, 20000] $ and $ j\in [50, 100] $, $ Y _{i, j} $ is novelset feature vector where $ i\in [1, 1000] $ and $ j\in [1, 64] $, where j=1 represents the original face feature vector and j=2,3,4,... represents SVD degraded face feature vectors for novelset.

For test set, we only focus on novelset persons, but search for the personIDs in both Baseset and Novelset. After dlib face feature extractor we finally got 4,899 face feature vectors although totally there are 5,000 face images in test set(some images may not detect any face with dlib).

Easiest, we just do a brute force search over all face feature vectors of 21,000 person in baseset and novelset(SVD degrade features are not included), search for the nearest face in totally 1,169,166 face images. And then we search for the most similarity personID for test set in way of compare with all feature vectors, similarity score defined bellow, and maximum similarity score return the particular personID. After a truly time consuming search, the final result is 13.39\%@P99, 33.41\%@P97, 56.87\%@P95.
\begin{equation}
SimilarityScore=1.0/(1.0 + dist(X, Y))
\end{equation}
\begin{equation}
dist(X, Y)=\sqrt{\sum_{i=1}^{n}(x_{i}-y_{i})^{^{2}}}
\end{equation}

Then, when SVD degrade features were included, and the result is 19.25\%@P99, 39.64\%@P97, 72.69\%@P95, which emphasize that SVD is truly useful for SIPP face recognition. Theoretically, when SVD face feature vectors included, it sames like that we added some intra-class variations for person in novelset, or more precisely, we have expanded high dimension feature space for novelset face image from on point to a hypersphere around the point. Showing in figure 3, in right circle, expanded from blue point to blue circle. It's really not a good result and we began to optimize the search methods and we get the evaluation result in table 1.

\begin{table*}
\begin{center}
\begin{tabular}{|c|c|c|c|}
\hline
Methods & P99 & P97 & P95\\
\hline\hline
Baseline (Base0) & 13.39\% & 33.41\% & 56.87\% \\
SVD degraded (svd+method1) & 19.25\% & 39.64\% & 72.69\% \\
Mean search (svd+method2) & 29.94\% & 92.65\% & \textbf {100\%} \\
Mean search+Brute-force (svd+method3) & 42.11\% & 93.55\% & \textbf {100\%} \\
\textbf {Mean serach+LSH (svd+method4)} & \textbf {47.52\%} & \textbf {94.2\%} & \textbf {100\%} \\
\hline
\end{tabular}
\end{center}
\caption{Coverage for all methods we used. Base represent brute force search over no SVD added feature vectors. At the same time, precision on baseset are greater than 98\% in all above evaluations.}
\end{table*}

As we could see, coverage has improved to 47.52\% after 4 methods added on base method superposed, and coverage reach to 100\%(P95) when only 2 methods had added, that is to say, when a precision of 95\% ensured, we can recognize all the people in test set. Next, we will introduce each of the methods and its theoretical explaining followed.

\begin{figure}[!htbp]
\center{\includegraphics[width=5.5cm]  {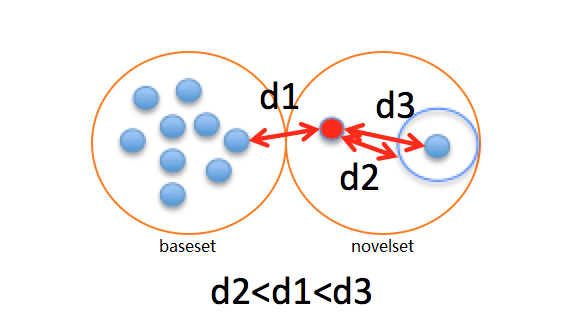}}
\caption{Explaining of why SVD face useful. Left circle: face from baseset which has many face feature vectors full-filed corresponding hyperspace. Right circle: face from novelset which has only one face feature vector and expanded after SVD faces added. Suppose red point is searching vector, and ground truth is red point belongs to personID of right circle, before SVD expanded, red point mis-classificated to left circle because d1\textless{d3} was found, after expanded, d2(d2\textless{d1}) could be found in order not to be mis-classified.}
\end{figure}

\begin{figure*}[!htbp]
\center{\includegraphics[width=15cm]  {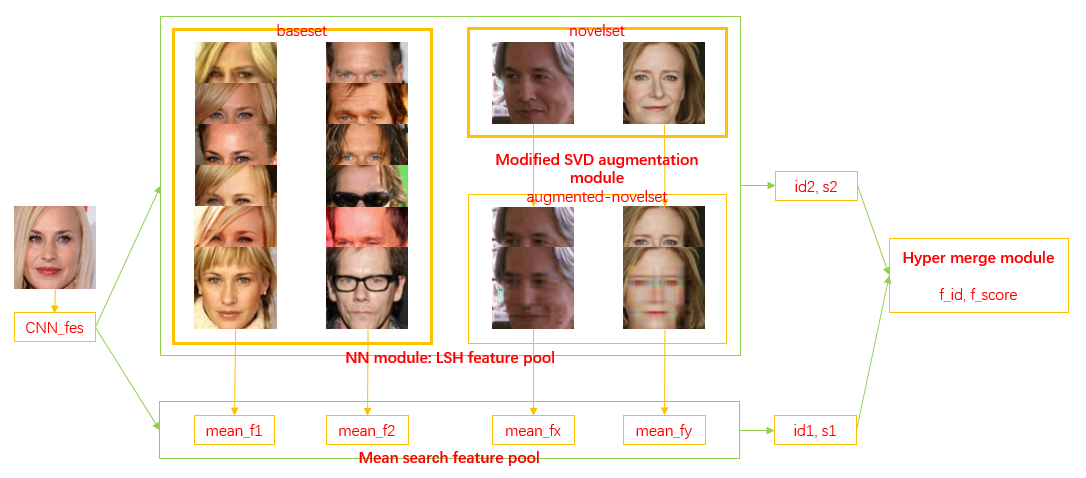}}
\caption{Pipeline of proposed method. Modified SVD augmentation module explained in 3.1, Mean search and LSH explained in 4.2, 4.4, Hyper merge module shown in 4.3 and equation 7.}
\end{figure*}
\section{Methods for searching the personID}
In this paper, we do not trying to training a deep neural networks that can have a good representation of each person with minimum intra-class and maximum inter-class variations. We focus on the search strategy for finding the corresponding personID. That means when a face feature extractor fixed, we still have many methods to improve the coverage and precision in SIPP face recognition problem. \textbf {Emphasize here, the combination of modified mean search and LSH method would be naturally leaded to step by step.} The overall pipeline shown in figure 4.

\subsection{Brute-force search with SVD degraded face}
As we have introduced in section 3.2, SVD face with 95\%, 90\%, 85\% energy reserved were included in the search space. On on hand, it's a way for us to added intra-class variations for persons in novelset. On the other hand, SVD added faces is also a way for us to expanding the representation of a person from one point to one hyperspace surrounding the point, although the hyperspace is not in the middle of the person feature space. This situation was explained in Figure 3.

After a heavy analysis on the evaluation, we found it's easy for us to find a person whose expression, light conditions, point of view and so on is quite similar to the search face but comes from another personID. In fact, this situation can also to be explained with Figure 3, although the hyperspace of novelset person has been expanded with SVD faces, but the expanding is limited because of SVD faces could only add variations in one way or in other words, the expand hyperspace is limited. How about we just utilize of mean feature vector to represent a person.

\subsection{Mean search with SVD degraded face}
Brute-force search is time-consuming, and it's directly we move to mean search, but there is a problem, faces in novelset has only one image and its mean vector is it's self or in other word, the mean do not add any more information. For person in novelset, mean vector was computed over all SVD degraded feature vectors. That means, only the main information of a person should be utilized, in order not to be misguided by variations such as expressions, light conditions, point of view which is not the essential difference between persons. But there is still drawbacks, for person in novelset, the mean feature vector is not in the middle of the high dimensions hyperspace of the person, but just located on the only one face feature vector we have. As the only one face feature vector is a little far from the middle of hyperspace of the person, it will always lead to not very high similarity score although its within the same person.
\begin{figure}[t]
\center{\includegraphics[width=5.5cm]  {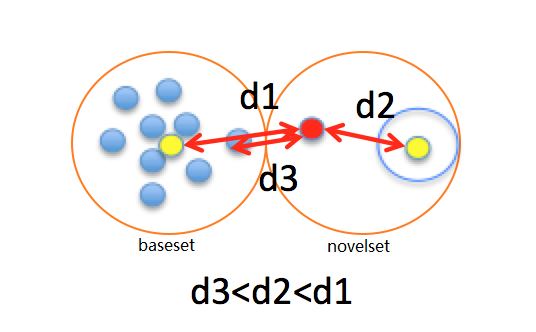}}
\caption{Explaining of why mean face feature vector useful. Red point is searching face, and left circle represent one person faces from baseset, right circle represent one person face from novelset. When brute-force search, d3(d3\textless{d2}) is the most similar one, but it's a mis-classification. When we just search over mean vector, only d1(d2\textless{d1}) would be founded, which would not lead to a mis-classified.}
\end{figure}

After mean vector search, it's clear that we've found an effective way to search for personID over baseset and novelset, as a coverage of 100\% has reached for precision 95\%(P95). So we keep mean search all the way with all methods we had next. Is brute-force search does not work at all? We do not think so. In fact, brute-force search can help to get a more distinguishable similarity score when two faces belongs to one person. So we just move forward one step, do a combination of mean search and brute force search.

\subsection{Combination of modified mean search and brute-force search}
As explained in method 4.2, mean search is a way for us to ensure a minimum coverage and precision in challenge-2, and brute force search is a way for us to get a more distinguishable similarity score which could increase the coverage a further step at a high precision(P99) required. In this part, an example why a more distinguishable similarity score would help and how we do would be introduced.

\begin{table}
\begin{center}
\begin{tabular}{|c|c|c|}
\hline
PersonId & SameIdNot & SimilarityScore\\
\hline\hline
0 & 1 & 0.92 \\
1 & 1 & 0.91 \\
2 & 1 & 0.90 \\
3 & 1 & 0.89 \\
4 & 1 & 0.88 \\
5 & -1 & 0.87 \\
{\textbf {6}} & {\textbf {1}} & {\textbf {0.86} $\uparrow$} \\
7 & -1 & 0.85 $\downarrow$ \\
{\textbf {8}} & {\textbf {1}} & {\textbf {0.84} $\uparrow$} \\
9 & -1 & 0.83 $\downarrow$ \\
\hline
\end{tabular}
\end{center}
\caption{Explain why a distinguishable similarity score is important. If we want to have a precision 100\%, the threshold will be 0.88 while coverage is 50\%. As personID 6 and 8 has a low similarity score of same person, If we can assign them a more distinguishable similarity score bigger than 0.88, or at the same time, we assign a lower score for personID 7 and 9, coverage would increase.}
\end{table}

\newcounter{firstlongequation}
\begin{figure*}
\normalsize
\setcounter{firstlongequation}{\value{equation}}
\setcounter{equation}{6}

\begin{equation}
SimilarityScore,Id=\left\{\begin{matrix}
max(s1,s2),id1 & if(id1=id2)\\
\left\{\begin{matrix}
s2,id2 & if(s2>s1+T)\\
s1,id1 & if(s2<s1-T)\\
min(s1,s2),id1 & if(s2\in [s1-T,s1+T])
\end{matrix}\right. & if(id1\neq id2)\\
\end{matrix}\right.
\end{equation}

\setcounter{equation}{\value{firstlongequation}}
\end{figure*}

In table 2, as the explaining, if we can assign a more distinguishable similarity score for which two face came from the same person with a higher confidence, coverage at a particular precision could be improved. If we want to get a higher similarity score, a brute force search could help in the way of d2 in figure 3. Also, the same with situation in figure 5, when do mean search, d3 would not be choose because of a mis-classification would occur. After a heavily analysis on evaluations, if we can find a face feature vector with a higher similarity score which exceed the similarity score of mean feature vector with a threshold T, that would be a important indication of same person, and if not, they always came from two persons. Based on this observation, we do a combination of mean search and brute force search in a rule based manner. Suppose after mean search, we found a most similar similarity score s1 and the corresponding personID id1, after a brute force search, another most similar similarity score s2 and the corresponding personID id2. The combination of mean search and brute-force search was done with rules explained in equation 7.

If mean search and brute force search find the same personID, we choose the maximum score between mean search and brute force search because we have a high confidence of a correct search. If mean search and brute force search indicate different personIDs, we choose the one who exceed the other with a score more than a threshold T. If the limitation of threshold T could not satisfied, we choose the minimum score between s1 and s2, and the mean search id for mean search has a higher confidence than brute force search in some extent. In this paper, threshold T is set to 0.03 for all evaluations.

\subsection{Combination of modified mean search and LSH}
On one hand, a brute force search method is extremely time consuming which could not been used in realtime applications. On the other hand, if noise exists in feature vectors, a brute force search method tends to mis-match with the noise as a higher similarity score may always been obtained at a constant probability. In order to handle the two weak points explained here, a nearest neighbor search method was used to do the combination with mean search. As nearest neighbor search always used in CBIR, K-D tress, structured K-means, LSH\cite{pauleve2010locality} are most commonly used methods. In this paper, a LSH based nearest neighbor search method was used and we set the target precision of search at 0.98 to ensure a good precision. As listed in table 1, we finally reach a coverage of 47.54\% at P99 and a coverage of 94.2\% at P97, 100\% at P95.

Each of the methods we conducted in section 4 has its intuitions behind. \textbf {SVD degraded face is a way for us to expand novelset faces from on point to a hypersphere}, brute-force search is direct but time-consuming and also could lead to much mis-classification in case of SIPP. \textbf {Mean search is a way for us to compute similarity with fundamental information of a face ignore its pose and expressions variations. Combination of brute-force search and mean search is a way to get a higher confidence score of a correct search. Combination of modified mean search and LSH is a way ignore noise in database}. In fact, hyper rules in equation 7 was inspired by table 2. If all mis-classified ones were given a lower similarity score and correct-classified were given a higher similarity score, coverage would improved.

\section{Evaluations and Analysis}
In this part, we shown more details for each evaluations in Msceleb challenge-2 dataset and CASIA WebFace dataset, which both were middle size face recognition dataset consists more than 10,000 persons and echo has 10-100 images per-person. In this paper, we do not evaluating in a normal manner, but formulate the evaluation in Msceleb challenge-2 2017 manner.

\begin{figure*}[!htbp]
\center{\includegraphics[width=10cm]  {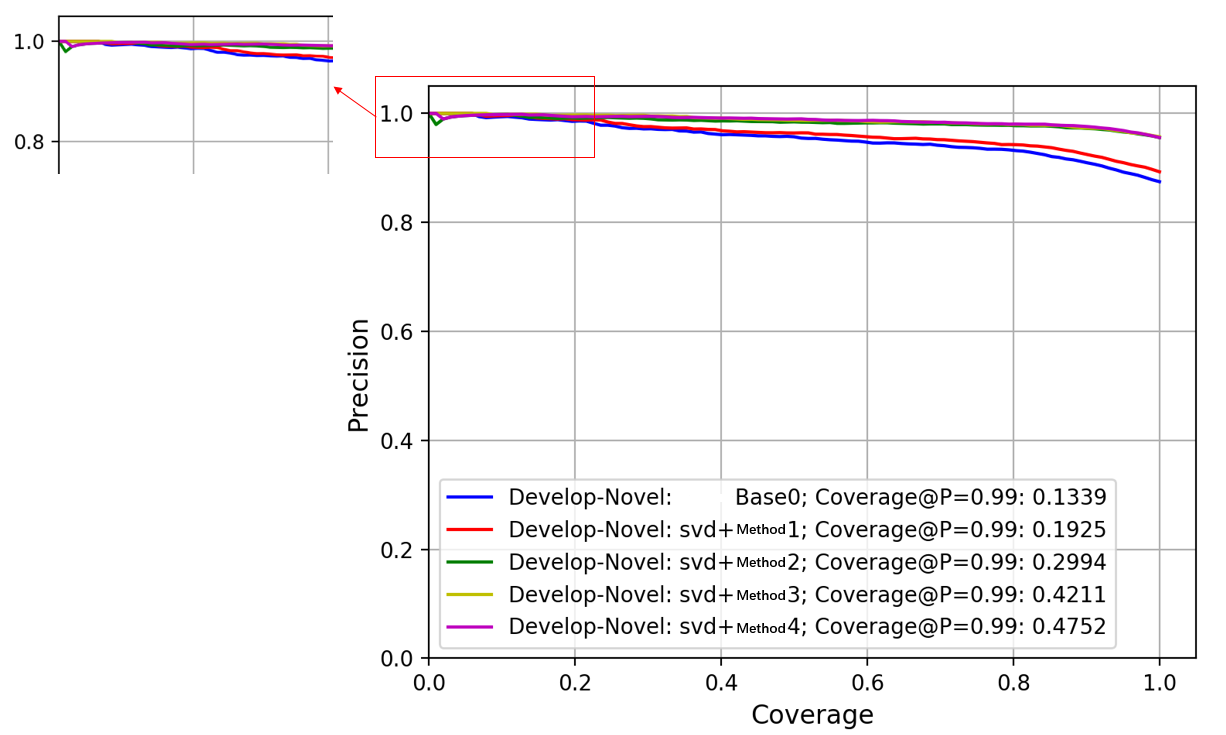}}
\caption{Precision Coverage @P99 for both sets. 5 methods were compared here, base0 is the brute-force search act as the baseline. Method 1-4 represent SVD degraded, Mean search, Mean search+Brute-force, Mean search+LSH respectively. Base0 and Method1 drops quickly at beginning of the curve means misclassification occurs even a higher confidence score given, which means the score is not reasonable. Method 2,3,4 solved the problem which means a higher coverage(or called recall), and curve always on top means a higher precision.}
\end{figure*}

\subsection{Evaluations on Msceleb dataset}
On Msceleb dataset, as shown in figure 6, precision-coverage curves was drawn. First, on the right side, when required precision is only 0.87, all 5 methods could reach a coverage of 100\%. Second, for method2, method3, method4, the final precision for all person in novelset is 95\%, which means the total number of mis-classified face images was same within the 3 methods. But, why it has a different coverage when precision is 99\%, the only explanation is that there is slightly different confidence score for some images, and that's why we are looking for a more reasonable way to assign a similarity score. Third, on the left side, base0 and method1 has a sharply drop at the beginning, that means we have assign a bigger similarity score for mis-matched faces or smaller similarity score for matched faces. After mean vector search added, this phenomenon disappeared, which also demonstrate the efficient of mean vector search.

Coverage over P97 and P95 are shown in figure 7 and figure 8. As coverage reaches 100\% when precision requirement is 95\%. We could also result in another conclusion, if a better face feature extractor would be obtained, which means the totally precision is higher than 95\%. In this way, we surly could get a higher coverage over P99.

\begin{figure}[t]
\center{\includegraphics[width=7cm]  {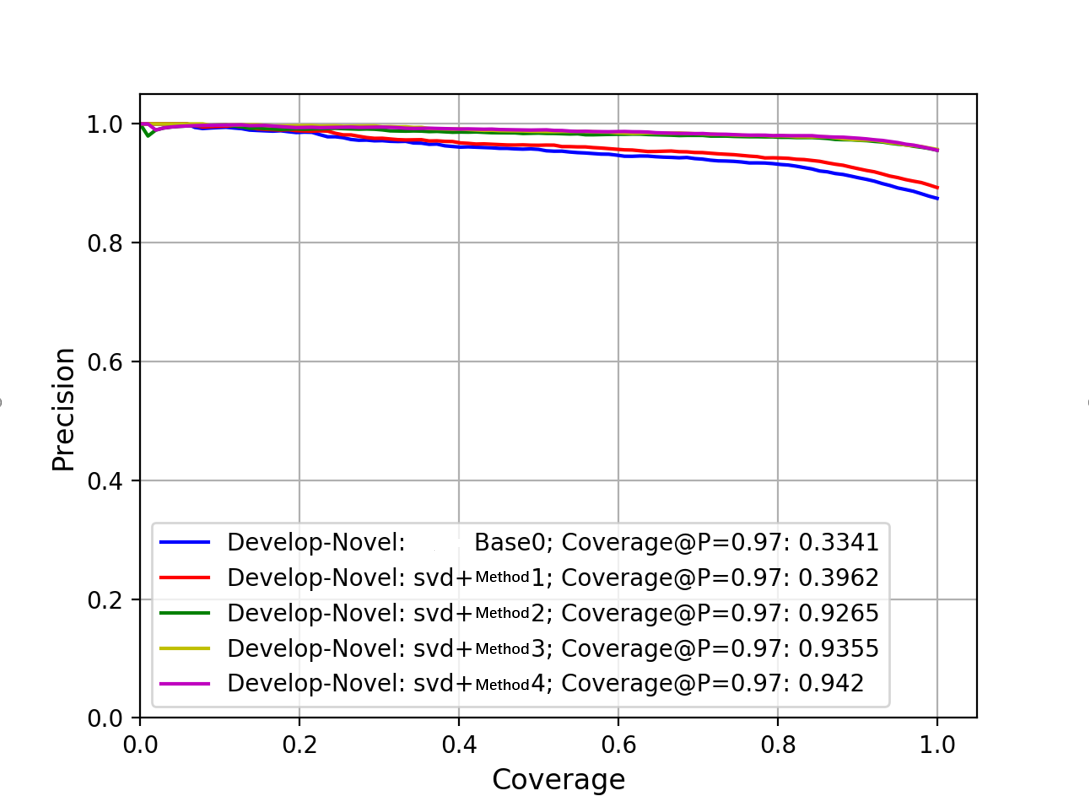}}
\caption{Precision Coverage @P97.}
\end{figure}
\begin{figure}[t]
\center{\includegraphics[width=7cm]  {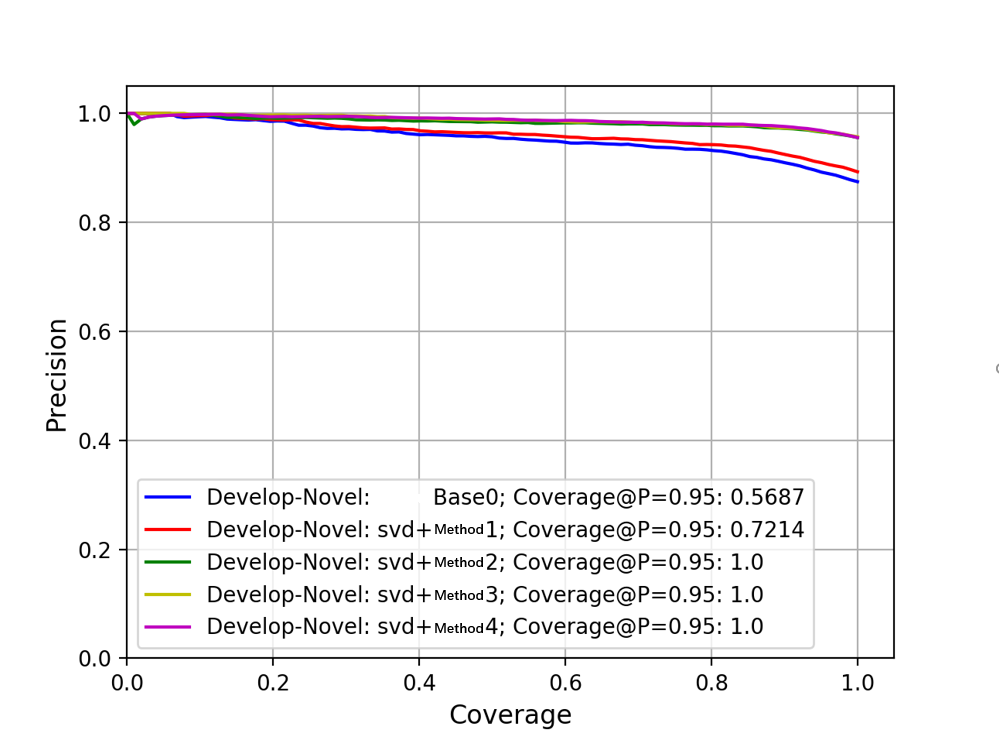}}
\caption{Precision Coverage @P95.}
\end{figure}

\begin{figure}[!htbp]
\center{\includegraphics[width=7cm]  {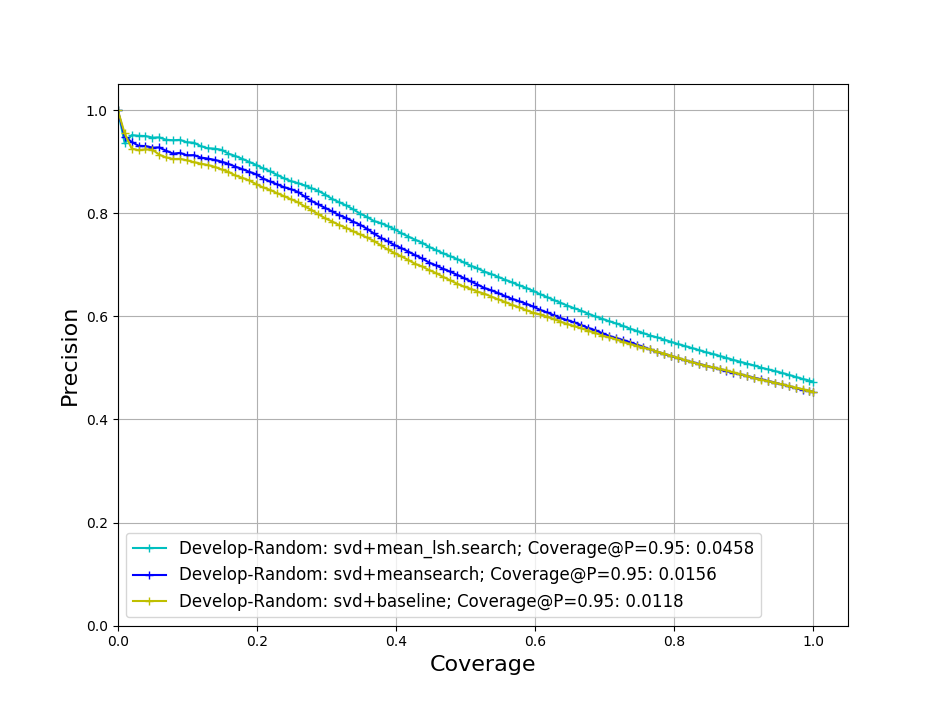}}
\caption{Precision Coverage @P95 Only for novelset. mean search+lsh is always on top, which indicating a higher recall and precision. But recall and precision is not high which indicating this is truly a hard task if we do not retrain DNN model.}
\end{figure}

\begin{table*}[!htbp]
\begin{center}
\begin{tabular}{|c|c|c|c|c|c|c|}
\hline
Methods & P98 & P97 & P95 & P90 & precision on novelset & precision on baseset\\
\hline\hline
SVD based brute-force search & 17.97\% & 29.15\% & 38.37\% & 50.41\% & 45.35\% & 86.42\%\\
Mean search & 21.17\% & 29.37\% & 39.11\% & 50.62\% & 45.37\% & 89.11\%\\
\textbf {Mean serach+LSH} & \textbf {21.38\%} & \textbf {32.19\%} & \textbf {41.44\%} & \textbf {52.43\%} & \textbf {47.35\%} & \textbf {89.38\%}\\
\hline
\end{tabular}
\end{center}
\caption{Coverage on WebFace dataset. Definition of P98, P97, P95, P90 are same with formal ones. precision on novelset and precision on baseset are separately precision of faces from novelset and baseset, but the testing are on both sets together.}
\end{table*}

\subsection{Evaluations on CASIA WebFace dataset}
On CASIA WebFace dataset, totally 494,414 images from 10,575 persons could be get public. For demonstrating our proposed methods in this paper, those person whose face images less than 10 are filtered in our testing, so 10,408 persons whose face images is bigger than 10 in our testing were used. Then we randomly selected nearly 10\% percent persons as novelset, and the rest baseset. For persons in baseset, only one random selected face were used for testing and the others for database, and for persons in novelset, only one random selected face were used for database and the others for testing. Here we just do comparison on SVD augmented brute-force search, mean search and the combination of mean search and LSH methods. Result list in Table 3 and Figure 9.

The same significantly improvements on novelset which we called SIPP also demonstrated the effectiveness of proposed method. We could see, the precision on baseset improved a little during each method, but the precision and recall improved much on novelset. As we could see, under P98, P97, P95 and P90, the coverage is really low, which also demonstrated that this task is really hard and should pay more attention to. Also, DNN model used in dlib (for feature extracting) also has a huge improvement space.

\subsection{Analysis}
In this paper, DNN model just act as a feature extractor and in fact other kinds of feature could also be used. In really world application, as explained in this paper, when new person registered, perhaps only one or two face images get, and it's not worth to retrain the DNN model(or some times, on embedded device, it's not allowed to retrain the model), methods in this paper could be used to improve the recall and precision on these novel persons. Intuitively, in face identification, when we searching all the face images dataset is time consuming, it also not a reasonable way for doing this, because mis-identify occurs when pose and expressions confused. Methods-4(combination of modified mean search and LSH search) is faster and more reliable, can be transferred without any extra efforts.

\section{Conclusions}
This paper do not aimed at design or training a powerful neural networks for SIPP face recognition problem, but pay more attention to proposing a combined search methods which would significantly increase the coverage under a given precision requirement with logic behind. First, a modified SVD based face image generate method was used to produce more intra-class variations. Second, 4 methods were proposed for more precise and effective search of correct personID and some theoretical explain followed. Third, we would like to emphasize, no need to retrain of the DNN model and this would be easy to be extended in many applications without much efforts. Coverage under P99 improved from 13.39\% to 47.52\% without retrain of model and much computing, and if we lower the precision requirement to P95, 100\% coverage obtained. Fourth, on evaluating on dataset CASIA WebFace dataset, the result also indicating that problems proposed in this paper is really a hard task and more attentions needed. In fact, this is truly the scenes in real applications: for some people, we could gathering a lots of there face images such as celebrity, and for other people, who are not famous and do not upload enough face images on internet, perhaps we could only get a identification card image, the proposed method would be very useful and easy to promoted. In the future, one on hand, we will focus on producing more intra-class variations for SIPP faces, with the way such as generative adversarial networks(GAN), one the other hand, more search methods could be evaluate for effective search with higher precision. Of course, we will try to design more powerful deep neural networks for a better face feature extractor to improve the performance of the whole system.

\bibliographystyle{ieee}
\bibliography{my-cv-texbib}

\end{document}